\documentclass[10pt,twocolumn]{article}

\usepackage{hyphenat}

\usepackage[linesnumbered,vlined]{algorithm2e}
\usepackage{booktabs} 
\usepackage{multirow}
\usepackage{multicol}
 \usepackage[table, svgnames, dvipsnames]{xcolor}
\usepackage{times}
\usepackage[pdftex]{graphicx}
\usepackage{epsfig}
\usepackage{mathtools}
\usepackage[pdftex]{geometry}
\usepackage{pdfpages}
\usepackage{indentfirst}
\usepackage[]{pdfpages}
\usepackage{booktabs}
\usepackage{multirow}
\usepackage{cite}
\usepackage[font=small,labelfont=bf]{caption}
\usepackage{hyperref}
\usepackage{dingbat}
\usepackage{bm}
\definecolor{kellygreen}{rgb}{0.5, 0.73, 0.09}
\hypersetup{
	pdfstartview={FitH},    
	pdfnewwindow=true,      
	colorlinks=true,       
	linkcolor=blue,          
	citecolor=blue,        
	filecolor=blue,      
	urlcolor=cyan           
}

\usepackage[num]{abntex2cite}
\usepackage{textcomp}

\usepackage{pgfgantt}
\usepackage{float}
\usepackage{sectsty}

\usepackage{titlesec}
\titleformat*{\section}{\large\bfseries}
\titleformat*{\subsection}{\normalsize\bfseries}
\titleformat*{\subsubsection}{\normalsize\bfseries}
\titleformat*{\paragraph}{\normalsize\bfseries}
\titleformat*{\subparagraph}{\normalsize\bfseries}
\titlespacing*{\section}
{0pt}{1ex plus 1ex minus .1ex}{1ex plus .1ex}
\titlespacing*{\subsection}
{0pt}{0.05ex plus 1ex minus .05ex}{0.05ex plus .05ex}

\usepackage{xcolor,colortbl}
\newcolumntype{g}{>{\columncolor{gray!20}}c}

\begin{document}
\twocolumn[
  \begin{@twocolumnfalse}
\begin{center}
\large{\textcolor{blue}{ 
This is a pre-print of an article published in {\bf Symposium on Applied Computing}. 
The final authenticated version is available online at: \url{https://doi.org/10.1145/3297280.3299728}
}}
\end{center}
\begin{center}
\Large{\bf BGrowth: an efficient approach for the segmentation of vertebral compression fractures in magnetic resonance imaging}
\end{center}
\begin{center}
	Jonathan S. Ramos$^{+}$\footnote{Corresponding author: jonathan@usp.br.}, 
Carolina Y. V. Watanabe$^{\dagger}$, \\
Marcello H. Nogueira-Barbosa$^{\star}$
and Agma J. M. Traina$^{+}$\\
\vspace{0.5cm}
$^{+}$Institute of Mathematics and Computer Science (ICMC), 
University of S\~ao Paulo (USP).
\\
$^{\dagger}$Computer Science Department (DCC), University of Rond\^onia (UNIR)\\
$^{\star}$Ribeir\~ao Preto Medical School (FMRP), 
University of S\~ao Paulo (USP). 
\end{center}

\begin{center}
\Large{\textbf{Abstract}}
\end{center}
\hrule
\vspace{0.2cm}
Segmentation of medical images is a critical issue: several process of analysis and classification rely on this segmentation. 
With the growing number of people presenting back pain and problems related to it, the automatic or semi-automatic segmentation of fractured vertebral bodies became a challenging task.
In general, those fractures present several regions with non-homogeneous intensities and the dark regions are quite similar to the structures nearby.
Aimed at overriding this challenge, in this paper we present a semi-automatic segmentation method, called Balanced Growth (BGrowth).
The experimental results on a dataset with 102 crushed and 89 normal vertebrae show that our approach significantly outperforms well-known methods from the literature.
We have achieved an accuracy up to 95\% while keeping acceptable processing time performance, that is equivalent to the state-of-the-art methods.
Moreover, BGrowth presents the best results even with a rough (sloppy) manual annotation (seed points).
\begin{center}
\end{center}
\hrule
\vspace{0.5cm}
  \end{@twocolumnfalse}
]

\section{Introduction}

Spinal diseases are quite usual worldwide and can cause significant loss of function and quality of life~\cite{pmid28472151}.
A very recurrent disease among older adults is the Vertebral Compression Fracture (VCF), which, in general, is caused by osteoporosis (benign) or bone metastasis (malignant)~\cite{pmid29422073,Barbieri2015,TEHRANZADEH2004440}. 
In general, VCFs are early detected or diagnosed based on shape or texture using Magnetic Resonance Imaging (MRI)~\cite{Antani2008,Paholpak2018,Barbieri2015,UETANI2004124}.
Usually, a specialist manually segments the Region of Interest (ROI) to aid the diagnose, which can be time consuming and prone to errors, due to inter and intra-subject variability and the subjective judgment that is employed~\cite{pmid29163946}.

\begin{figure}[!htb]
	\centering
\footnotesize
\setlength{\tabcolsep}{3pt} 
	\begin{tabular}{cc} 
		\includegraphics[width=0.44\linewidth]{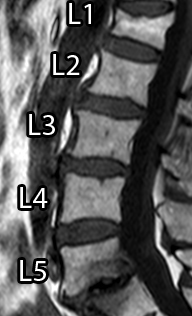} &
	    \includegraphics[width=0.44\linewidth]{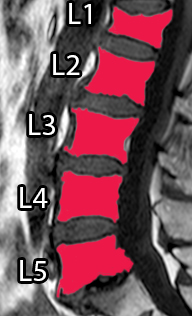} \\
			(a) Original image & (b)  Ground-truth \\
	\end{tabular}
		\caption{Example of manual vertebral bodies segmentation.}
	\label{fig:malignas}
\end{figure}

However, it incorporates expert knowledge gained over several years.
\autoref{fig:malignas} shows an example of manual segmentation over five lumbar vertebral bodies (L1-L5).
Computer-Aided Diagnosis (CAD) systems with automatic or semi-automatic segmentation methods are quite useful in this case, as long as they present fast results and delineation closer/similar to the manual segmentation~\cite{pmid29163946,casti2017}. 
Hence, an accurate segmentation algorithm plays a significant role and can assist the medical specialist in surgical planning and evaluation of suitable treatments~\cite{SpineWeb11}, for example, segmentation of Lung nodule~\cite{El-Baz2013,Soliman2018} and cortical bone~\cite{Hafri2016,Hafri2016b}.

The semi or automatic segmentation of VCFs is a challenging task, due to non-homogeneous gray-scale intensities within the same vertebral body (for example, L5 in \autoref{fig:malignas}).
To overcome this issue, several one-seed-point approaches have been proposed. In~\cite{Barbieri2015} is presented the VBSeg method, which employs superpixels, region growing and Otsu threshold. 
Region growing techniques are used in~\cite{casti2017}, such as, snakes (Chan-Vese), Otsu and fuzzy c-means clustering, in order to compose a cooperative strategy for a dynamic ensemble of classification models.
Although both works have presented very promising results, the VBSeg method presents a low segmentation performance on VCFs (61\% and 74\% Jaccard coefficient for malignant and benign VCFs, respectively).
The cooperative strategy for classification still depends on the ground-truth.

In \cite{pmid29163946}, the GrowCut~\cite{Vezhnevets05growcut} algorithm is employed for the segmentation of normal vertebral bodies.
The GrowCut method and its faster version, named as Fast GrowCut~\cite{FastGrowCut2014} (presents slightly lower segmentation performance than the original GrowCut~\cite{FastGrowCut2014}), employ several seeds points inside and outside the object of interest and have been widely used for many medical MRI exams (especially in oncology)~\cite{FERREIRAJUNIOR201823}.
However, to the best of our knowledge, GrowCut was not tested on VCFs.

Based on the formulation of segmentation as an energy minimization problem, many algorithms have been proposed, such as GrabCut (GB) and Lazy Snapping (LS).
GrabCut uses Gibbs energy~\cite{Geman1984} and Gaussian Mixture Models (GMM)  as soft segmentation for the background and foreground~\cite{Rother2004}.
LazySnapping works as an interactive image cutout tool and also uses Gibbs energy, combining graph cut with a pre-computed over-segmentation, e.g. Superpixels~\cite{LazySnapping2004}.
Although GB and LS have not been used for the segmentation of VCFs, they have been adapted or used in several medical applications~\cite{YONG201738,Lu2017,Wu2016}.
We consider both approaches in our analysis to address a wider assessment of VCFs segmentation over the state-of-the-art methods.

Aimed at overriding the challenge of VCFs segmentation, we propose the Balanced Growth (BGrowth) method, which balances the weights employed during the regions expansion.
The experimental results show that our approach  significantly outperforms the methods from the literature, achieving 95\% accuracy while keeping an processing time equivalent to the competitor methods. 
Moreover, we achieved the best results even with sloppy annotations.
%

The remainder of the paper is structured as follows. 
%
First, in \autoref{sec:proposed-method}, we describe our proposed approach for the segmentation of vertebral bodies in MRI.
Then, in \autoref{sec:materialsMethods}, we explore the materials and methods used in our work.
Next, in \autoref{sec:experiments}, we detail the experiment design, results and discussion.
Finally, the conclusions are presented in \autoref{sec:conclusion}.

\section{Balanced Growth: the proposed method}
\label{sec:proposed-method}
\begin{figure}[t]
	\centering
	\includegraphics[width=0.95\linewidth]{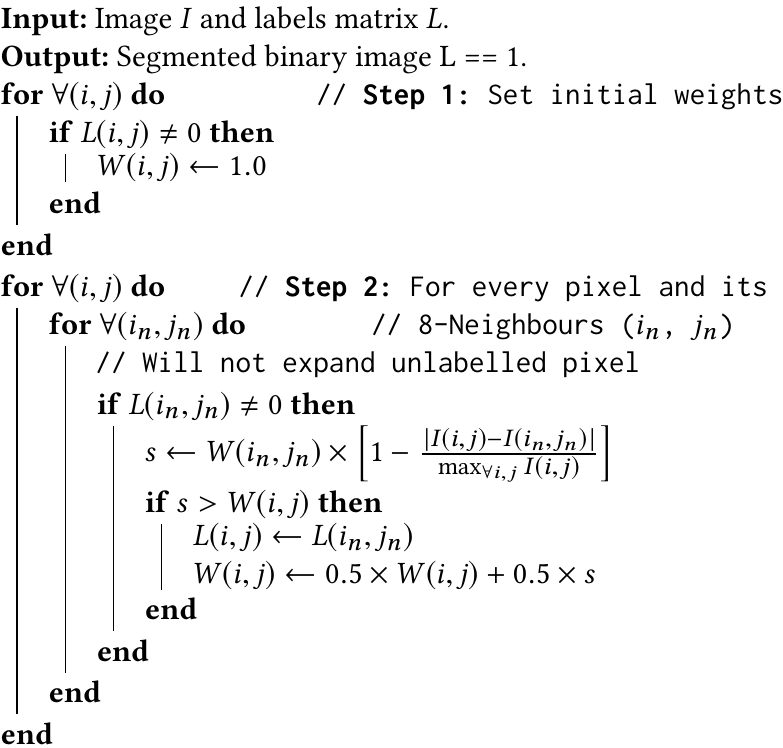}
    \caption{Balanced Growth (BGrowth) method.}
	\label{fig:gcPseudoCodeP}
\end{figure}

Our proposed approach is mainly based on the motto ``unity is strength''.
We take into account that all pixels in the path of an expanding region have influence on its growth.
We achieve ``unity'' by balancing the weights along the growing path, therefore the final segmentation is a ``unanimity'' among the pixels.
This allows small intensities transitions (as occurs mostly often in fractured vertebral bodies) to be better delineated by \textbf{B}alanced \textbf{Growth} (BGrowth).

For the sake of simplicity, we consider the segmentation of two regions (foreground and background) and gray-scale images.
However, the method can be easily adapted to two or more regions and for any color space.
Considering a digital image $I$ and its annotations or labels regions as a matrix $L$, both with dimension $M \times N$ ($M$ is the number of rows and $N$ the number of columns), \autoref{fig:gcPseudoCodeP} shows an overview of BGrowth.
Each entry in $L$ has value $-1$, $0$ or $1$, representing background, unlabelled and foreground, respectively.

The first step is to set up a weight matrix $W$ as follows
\begin{equation}
W(i,j) = \begin{cases}
$1.0$ &\text{if $L(i,j) \neq 0$}\\
$0.0$ &\text{otherwise} 
\end{cases}
\end{equation}
in which $i = 1, 2, ... , M$ and $j = 1,2,..., M$ are the matrices indexes and $W$ has the same dimension as $I$ and $L$.



    
            

The second step is to check each pixel and its 8-neighbours in the image.
For each neighbouring pixel at index $(i_n, j_n)$, a strength factor $s$ is calculated as
\begin{equation}
  s \leftarrow W(i_n,j_n) \times \left[1 - \frac{|I(i,j) - I(i_n, j_n)|}{\max_{\forall (i,j)} I(i,j)}\right] \ , 
\end{equation}
in which the absolute pixel intensity difference is normalized by the maximum pixel value in the image and shifted by 1, which produces values within the range $[0,1]$.
Then, the strength factor is the product of the neighbouring weight and the pixel difference aforementioned.
If $s$ is greater than the current pixel weight, then the neighbouring pixel expands to the pixel at index $(i,j)$: its label is changed and the new weight is balanced with the old one, i.e. $W(i,j) = 0.5\times W(i,j) + 0.5 \times s$.

\begin{figure}[t]
	\centering
	\footnotesize
		\setlength{\tabcolsep}{4pt} 
		\begin{tabular}{ccc}
		\includegraphics[width=0.3\linewidth]{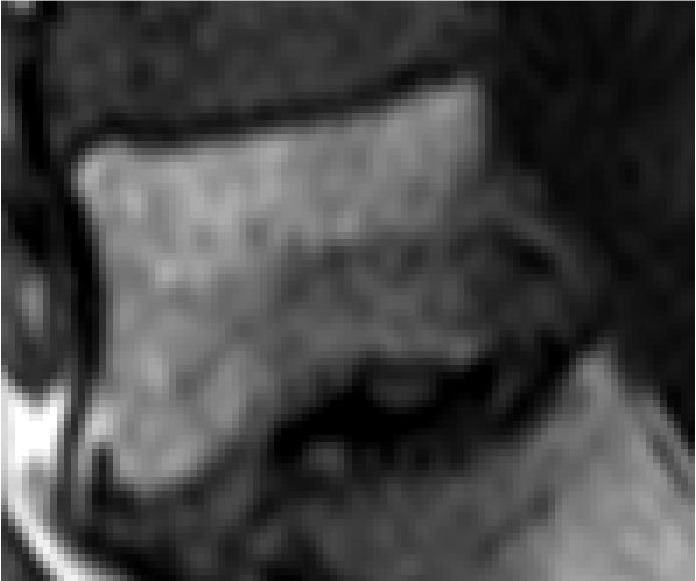}
	      & \includegraphics[width=0.3\linewidth]{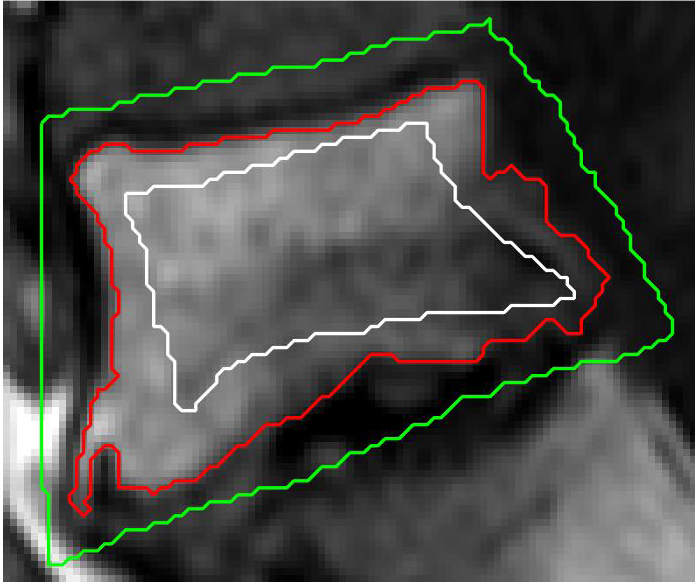}
	      &  \includegraphics[width=0.3\linewidth]{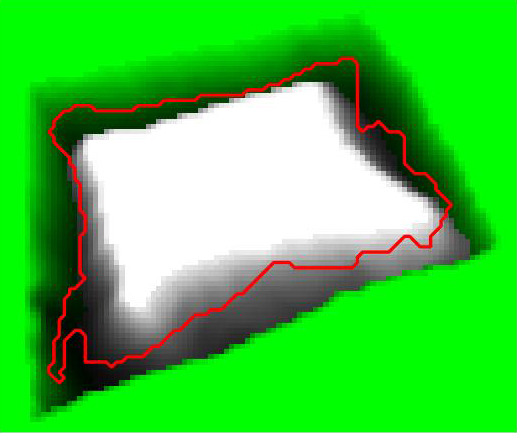}  \\
	       Original ROI & GT and seeds &  Iteration 1 \\

	      \includegraphics[width=0.3\linewidth]{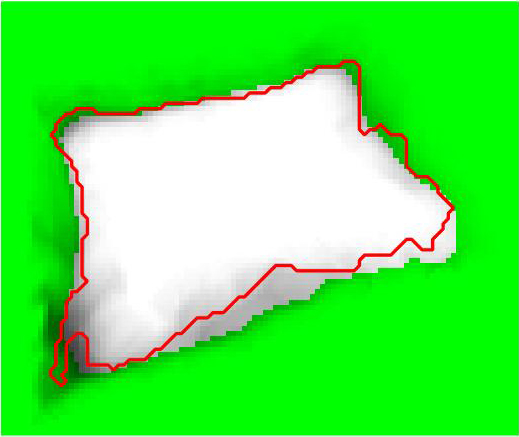} &
	      \includegraphics[width=0.3\linewidth]{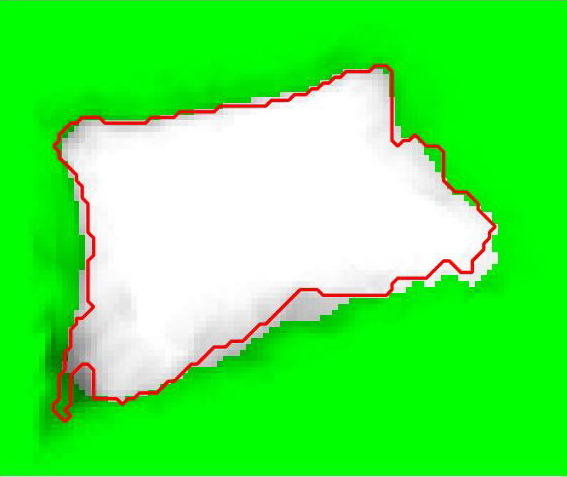}
	      & \includegraphics[width=0.3\linewidth]{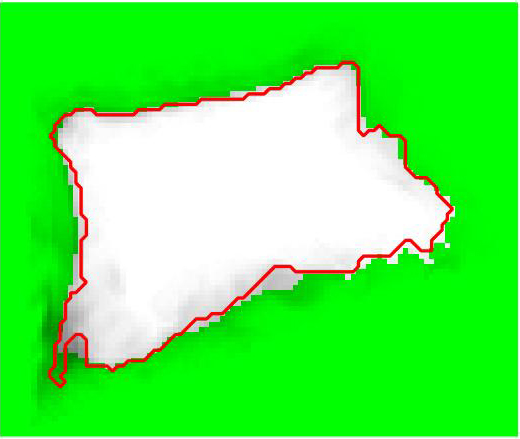}\\
	      Iteration 5 & Iteration 8 & Iteration 10 \\

	       \includegraphics[width=0.3\linewidth]{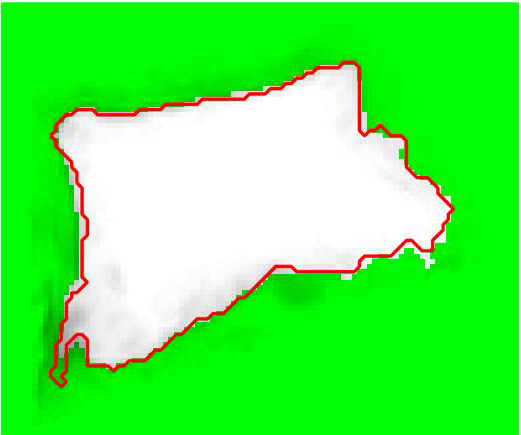} 
	      & \includegraphics[width=0.3\linewidth]{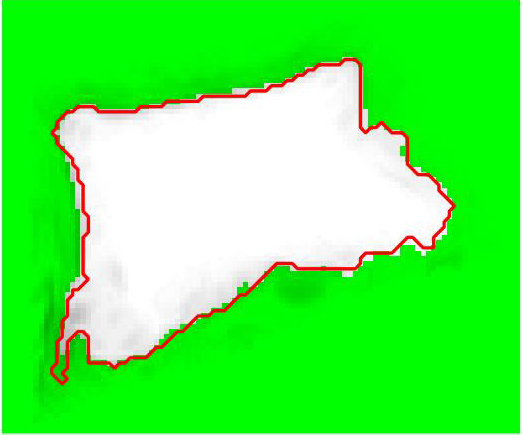}
	      & \includegraphics[width=0.3\linewidth]{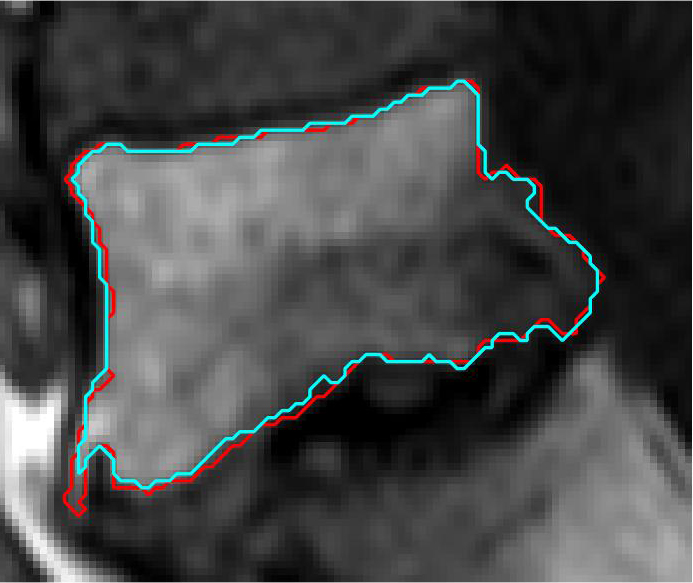}\\
	      Iteration 15 & Iteration 30 & Final result \\
	\end{tabular}
		\caption{BGrowth' iterations: Ground-Truth (GT), interior annotation, exterior annotation and final result boundaries are outlined in red, white, green and cyan, respectively.}
	\label{fig:GC-GCS-examples}
\end{figure}

\begin{figure}[t]
	\centering
	\footnotesize
		\setlength{\tabcolsep}{4pt} 
		\begin{tabular}{ccc}
		    \includegraphics[width=0.3\linewidth]{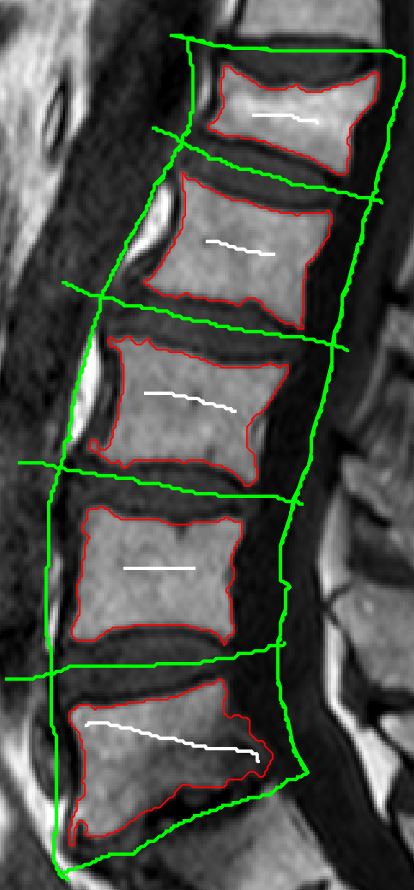} & \includegraphics[width=0.3\linewidth]{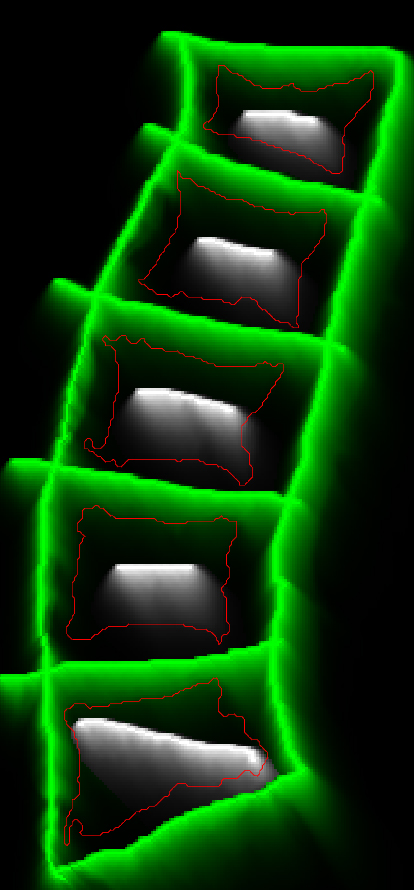} & \includegraphics[width=0.3\linewidth]{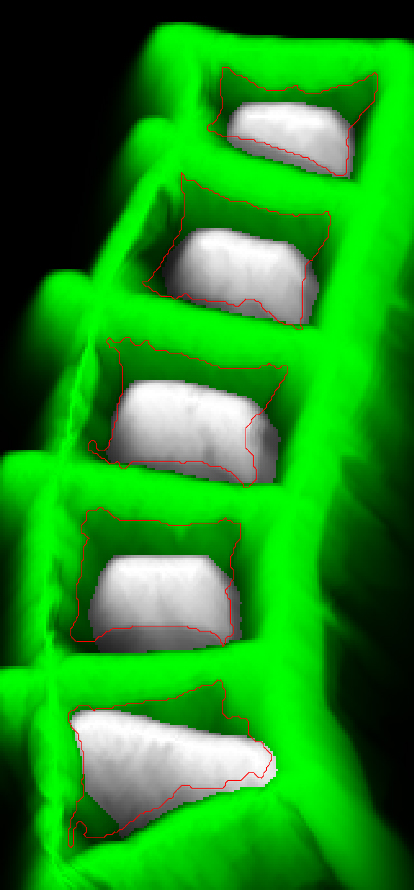} \\
		    Annotations & Iteration 1 &  Iteration 5 \\
	       \includegraphics[width=0.3\linewidth]{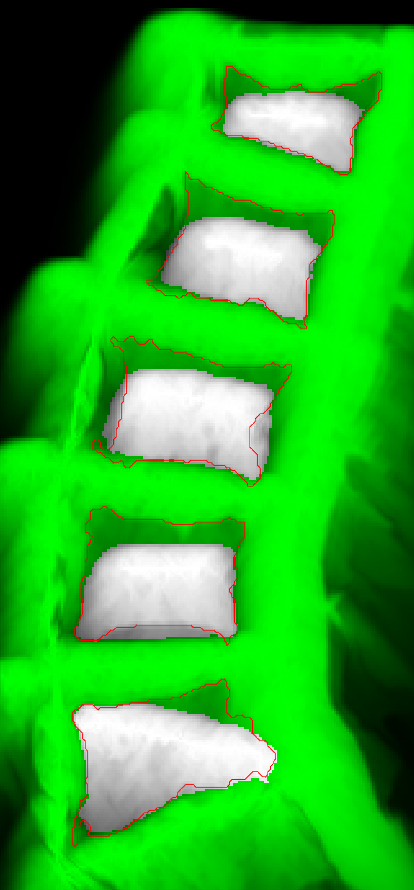} & \includegraphics[width=0.3\linewidth]{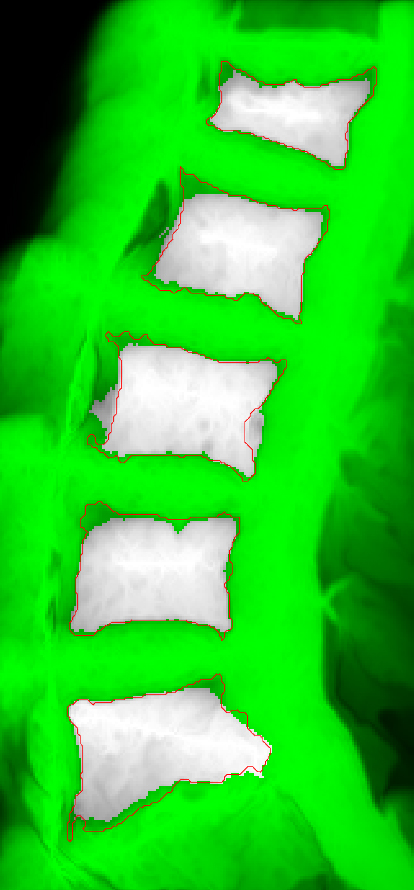}  & \includegraphics[width=0.3\linewidth]{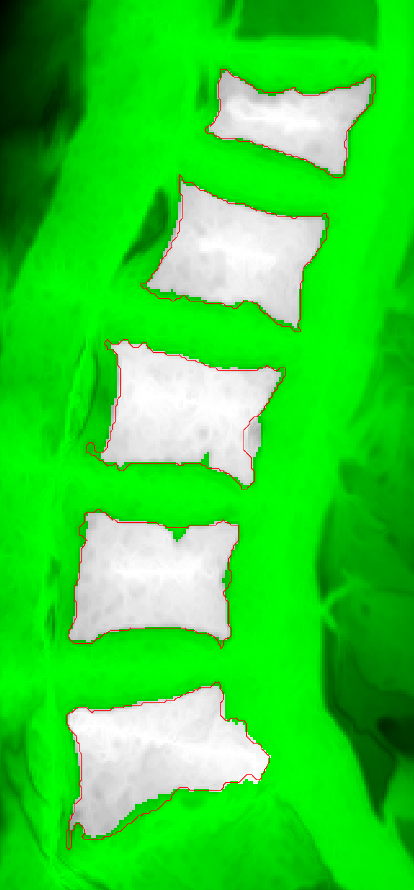}\\
	        Iteration 10 & Iteration 20 & Iteration 50 \\
	\end{tabular}
		\caption{BGrowth' iterations on five lumbar: Ground-Truth, interior annotation and exterior annotation boundaries are outlined in red, white and green, respectively.}
	\label{fig:BG-wholeLumbar}
\end{figure}

The balance induced by the weights produces smoother borders at regions with small pixel intensities.
Moreover, as long as the interior annotation contains small portions of the dark and bright regions, it can work well even with rough interior and exterior seed points.
To better illustrate this, \autoref{fig:GC-GCS-examples} shows a few iterations of BGrowth for a single vertebral body segmentation.
Note that, at iteration 5, a few dark regions (outside of the vertebral body) are still part of the foreground and, as the balancing goes on, the foreground shrinks towards the ground-truth.
The final result is quite close to the manual segmentation.
\autoref{fig:BG-wholeLumbar} shows another example of BGrowth's iterations on five lumbar vertebrae.
Note that, the method works quite well even when a simple line is given as seed points.
The only constraint is that the interior annotation has to comprise the dark/bright regions within the vertebral body.


As we work with the balance of old and new strength factors, it might happen that the process of expansion goes back and forth on the same pixel.
Therefore, we would recommend the definition of a maximum number of iterations depending on the kind of image being segmented.
We have empirically used a maximum of 30 iterations for the segmentation of a single vertebral body.
For a whole exam, as exemplified in \autoref{fig:BG-wholeLumbar}, a few more iterations might be necessary.

In the worst case scenario, the algorithm has a time complexity of $M \times N \times 8 \times it = O (M\times N\times it)$, where  $it$ is the maximum number of iterations defined.

\section{Materials and methods}
\label{sec:materialsMethods}
In this section, we first describe the experimental dataset, which is composed of normal, malignant and benign vertebral bodies. 
Then, we present the comparison measures used.
Next, we report the segmentation methods and parameters settings.
Finally, the annotation scheme and computation set-up are explored.

\subsection{Image dataset}

Due to space limitations, we will show the results obtained from only one meaningful dataset of lumbar vertebral bodies (L1, L2, L3, L4, and L5), including 89 normal vertebral bodies and 102 with VCFs. 
Bone metastases occurred in 49 VCFs (malignant VCFs) and were confirmed by biopsy and histopathological analysis.
The remaining 53 VCFs were benign fractures and, following clinical guidelines, not all of them were confirmed by biopsy or histopathological analysis.
On these cases, a musculoskeletal radiologist with over 20 years of experience carefully reviewed the radiological information system (RIS) and the hospital information system (HIS) and supervised the manual segmentation (ground-truth). This study was approved by the Ethics Research Committee of the Ribeir\~ao Preto Medical School - USP, where the dataset was acquired.

\subsection{Comparison measures}
We analyzed the Jaccard Coefficient $J$ and Dice Score $D$~\cite{Jaccard12similarityCoefficient,sorensen1948method}:
\begin{equation}
     J(GT,Seg) = \frac{|GT \cap Seg|}{|GT \cup Seg|} \ ; 
    \label{eq:jaccard1}
\end{equation}
\begin{equation}
     D(GT,Seg) = \frac{2\times |GT \cap Seg|}{|GT| + |Seg|} \ , 
    \label{eq:dice1}
\end{equation}
in which $GT$ represents the ground-truth region and $Seg$ represents the region yielded by the segmentation technique.

For further analysis, we also employed the measures of accuracy ($A$) precision ($P$) and recall ($R$)~\cite{RAMOS2017,Ramos2016}:
\begin{equation}
\textnormal{A} = \frac{TP+TN}{TP+FP+TN+FN} \ ; 
\label{eq:accuracy}
\end{equation}
\begin{equation}
\textnormal{P} = \frac{TP}{TP+FP} \ \ ; \ \ \textnormal{R}  =  \frac{TP}{TP+FN} \ ,
\label{eq:precision}
\end{equation}
in which \textbf{True Positive (TP):} number of pixels correctly segmented as part of the vertebral body; \textbf{True Negative (TN):} number of pixels correctly segmented as part of the background; \textbf{False Positive (FP):} total of pixels miss-segmented as belonging to the vertebral body; \textbf{False Negative (FN):} number of pixels belonging to the vertebral body miss-segmented as part of the background.

The precision measures the percentage of pixels correctly segmented, considering the FPs.
The higher the number of FPs, the lower the precision, indicating how much the method segments outside of he ground-truth.
Likewise, the recall measures the percentage of pixels correctly segmented considering the FNs.
The higher the number of FNs, the lower the recall, indicating how much is not segmented on the inside of the vertebral body.

Another measure that quantifies the balance between P and R is the F-measure, defined as~\cite{Goutte2005} 
\begin{equation}
\textnormal{F} = 2 \times \frac{P \times R}{P + R}.
\label{eq:f-measure}
\end{equation}

\autoref{tb:descriptions} shows a summary of the segmentation methods and comparison measures used in this work.
\begin{table}[t]
\centering
	\setlength{\tabcolsep}{4pt} 
\footnotesize
\caption{Summary of Acronyms/symbols used in this work.}
\label{tb:descriptions}
\begin{tabular}{rlrlc}
\toprule
\multicolumn{2}{c}{Segmentation Methods} & \multicolumn{2}{c}{Comparison Measures}\\
Abbr. & Description&  Symb. & Description\\ \midrule \midrule
 BG &  Balanced Growth & A & Accuracy\\
 CV & Chan-Vese & $D$ & Dice-Score\\
 GB  & GrabCut & F & F-measure \\
 GC &  GrowCut & $J$ & Jaccard coefficient\\
 LS & LazySnapping & P & Precision\\
 OT & Otsu & R &  Recall\\
\bottomrule
\end{tabular}
\end{table}

\subsection{Segmentation algorithms}
In order to evaluate the performance of BGrowth, we compared it with the methods\footnote{Deep learning approaches were not used due to the small number of ROIs available.} GrowCut (GC), GrabCut (GB), LazySnapping (LS), VBSeg (VBS), Snakes (Chan-Vese, CV) and Otsu threshold (OT).

\subsection{Parameters and settings}
The parameters for all methods that used superpixels were empirically set to $m\times n \times 0.25$  superpixels (25\% of the total pixels in the image) for each ROI, in which $m$ and $n$ are the number of rows and columns of the ROI, respectively.
The maximum number of iteration was set to 300 for Chan-Vese technique and 30 to GrowCut and BGrowth.
The remainder of the parameters for all methods were set to default settings to avoid loss of generalizability.
No pre or post-processing technique were applied to assure the same conditions for all segmentation methods.

\subsection{Annotations}
As \autoref{fig:an-examples} shows, the initial interior and exterior seeds annotation were performed in a ``sloppy'' way, i.e., no detailed boundary for accentuated curves were drawn.
In general, the annotation looks like a rectangle.

\subsection{Computational set-up}

The experiments were performed on a desktop with a 3.60GHz Intel(R) Core(TM) i7 CPU and 16GB RAM using Matlab(R) version 2018a.

\section{Experiments, results and discussion}
\label{sec:experiments}

The experiments are analyzed in three main parts.
First we discuss results for the six measures aforementioned, then we analyzed their statistical difference for each segmentation method against Balanced Growth (BGrowth or BG).  
As a final analysis, we assess the impact of changing the percentage of manual annotation inside and outside each vertebral body for every approach that uses this kind of annotation.

\subsection{Overall measures analysis}

\autoref{tb:AJD} reports the average Accuracy (A), Jaccard coefficient (J) and Dice score (D) achieved by each segmentation method in our experiments separated by cases `All', `Normal', `Benign' and `Malignant'. 
Note that, for A, J and D: 
\begin{itemize}
    \item BGrowth (BG) presented the best results for all cases; 
    \item LazySnapping (LS) presented the second best results for cases  `All', `Normal' and `Benign';
    \item  GrowCut (GC) presented the second best results for `Malignant' case;
    \item  All the others methods presented results equal or below 80\%; 
\end{itemize}
\begin{figure}[t]
	\centering	
\footnotesize
		\setlength{\tabcolsep}{0.5pt} 
	\begin{tabular}{cc}
	     \includegraphics[width=0.49\linewidth]{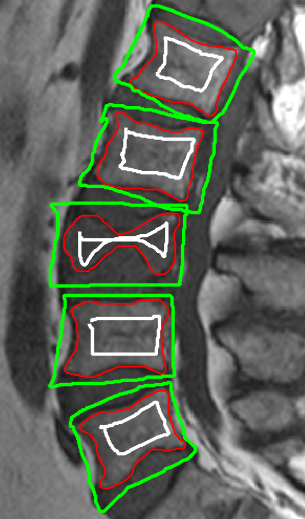} &  
	     \includegraphics[width=0.49\linewidth]{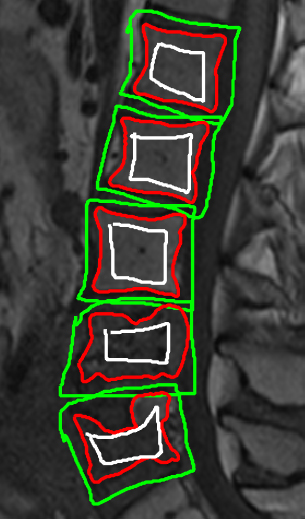} \\
	\end{tabular}
		\caption{Example of sloppy annotations (seed points): ground-truth (GT) in red; interior and exterior annotations in white and green, respectively.}
	\label{fig:an-examples}
\end{figure}
\begin{table}[!bht]
\centering	
\footnotesize
		\setlength{\tabcolsep}{4pt} 
  \caption{Comparison of Accuracy (A), Jaccard Coefficient (J) and Dice Score (D) for all methods considering distinct cases. The best value is in \textbf{bold} and the second one is underlined.} 
  \label{tb:AJD}
\begin{tabular}{rcccccc}  
\toprule
 & \multicolumn{6}{c}{Vertebral bodies considered (\%)} \\ \cmidrule{2-7}
& \multicolumn{3}{c}{All} & \multicolumn{3}{c}{Normal} \\
 \multirow{-3}{*}{\rotatebox[origin=c]{90}{Method}} & A & J & D & A & J & D  \\ \midrule \midrule
 BG &  \textbf{95$\pm$05} & \textbf{85$\pm$16} & \textbf{91$\pm$15} & \textbf{96$\pm$06} & \textbf{88$\pm$18} & \textbf{92$\pm$19} \\
 
\rowcolor{gray!10} GC & 93$\pm$05 & 81$\pm$15 & 88$\pm$15 & 94$\pm$06 & 85$\pm$18 & 90$\pm$18 \\
 
LS & \underline{94$\pm$05} & \underline{84$\pm$16} & \underline{90$\pm$15} & \underline{95$\pm$05} & \underline{87$\pm$18} & \underline{91$\pm$19} \\
 
\rowcolor{gray!10}GB & 76$\pm$12 & 58$\pm$16 & 72$\pm$16 & 78$\pm$11 & 62$\pm$17 & 75$\pm$17 \\
 
CV & 75$\pm$21 & 59$\pm$22 & 72$\pm$20 & 84$\pm$15 & 70$\pm$19 & 80$\pm$19 \\
 
\rowcolor{gray!10}OT & 78$\pm$13 & 53$\pm$26 & 64$\pm$28 & 84$\pm$09 & 66$\pm$21 & 76$\pm$22 \\
 
  VBS & 71$\pm$110 & -- & -- & 
-- & -- & -- \\ \midrule 

& \multicolumn{3}{c}{Benign} & \multicolumn{3}{c}{Malignant}  \\  \midrule \midrule
BG & \textbf{95$\pm$05} & \textbf{83$\pm$15} & \textbf{90$\pm$14} & \textbf{92$\pm$02} & \textbf{80$\pm$07} & \textbf{89$\pm$04}\\ 
\rowcolor{gray!10}GC& 92$\pm$05 & 76$\pm$14 & 85$\pm$13 & \underline{92$\pm$03} & \underline{79$\pm$07} & \underline{88$\pm$04}\\ 
LS & \underline{94$\pm$05} & \underline{82$\pm$15} & \underline{89$\pm$14} & 92$\pm$04 & 79$\pm$09 & 88$\pm$06\\ 
\rowcolor{gray!10}GB & 72$\pm$11 & 52$\pm$16 & 67$\pm$15 & 75$\pm$12 & 58$\pm$13 & 72$\pm$10\\ 
CV & 76$\pm$22 & 57$\pm$23 & 69$\pm$21 & 55$\pm$18 & 41$\pm$12 & 57$\pm$12\\ 
\rowcolor{gray!10}OT  & 81$\pm$11 & 53$\pm$23 & 66$\pm$23 & 64$\pm$11 & 26$\pm$19 & 38$\pm$25\\
VBS  & 75$\pm$--- & -- & -- & 
62$\pm$--- & -- & --\\
 \bottomrule
\end{tabular}
\end{table}
\begin{table}[!bht]
\centering	
\footnotesize
		\setlength{\tabcolsep}{4pt} 
  \caption{Comparison of the Precision (P), Recall (R) and F-Measure (F) for all methods considering distinct cases. The best value is in \textbf{bold} and the second one is underlined.} 
  \label{tb:AJD2}
  
\begin{tabular}{rcccccc}  

\toprule
 & \multicolumn{6}{c}{Vertebral bodies considered (\%)} \\ \cmidrule{2-7}
& \multicolumn{3}{c}{All} & \multicolumn{3}{c}{Normal}   \\ 
  \multirow{-3}{*}{\rotatebox[origin=c]{90}{Method}} & P & R & F & P & R & F  \\ \midrule \midrule
 BG & \underline{88$\pm$16} & \underline{94$\pm$14} & \textbf{91$\pm$15} & \underline{91$\pm$19} & \underline{93$\pm$19} & \textbf{92$\pm$19} \\ 
 
 \rowcolor{gray!10}GC & 84$\pm$16 & 94$\pm$15 & 88$\pm$15 & 88$\pm$18 & 92$\pm$19 & 90$\pm$18\\
 
 
 LS & \textbf{90$\pm$17} & 91$\pm$15 & \underline{90$\pm$15} & \textbf{92$\pm$19} & 90$\pm$19 & \underline{91$\pm$19} \\ 
 
\rowcolor{gray!10} GB & 59$\pm$18 & \textbf{96$\pm$16} & 72$\pm$16 & 63$\pm$18 & \textbf{95$\pm$20} & 75$\pm$17 \\ 
 
 CV & 66$\pm$24 & 88$\pm$17 & 72$\pm$20 & 75$\pm$21 & 90$\pm$16 & 80$\pm$19 \\ 
 
\rowcolor{gray!10} OT & 61$\pm$26 & 72$\pm$31 & 64$\pm$28 & 70$\pm$22 & 86$\pm$22 & 76$\pm$22\\

   VBS &  80$\pm$140 & 87$\pm$060 & -- & 
 -- & -- & --  \\ \midrule 
 
 & \multicolumn{3}{c}{Benign} & \multicolumn{3}{c}{Malignant} \\ \midrule \midrule
BG & \underline{87$\pm$16} & \underline{94$\pm$12} & \textbf{90$\pm$14} & \underline{83$\pm$07} & \underline{96$\pm$03} & \textbf{89$\pm$04} \\
\rowcolor{gray!10}GC  & 79$\pm$15 & 94$\pm$13 & 85$\pm$13 & 82$\pm$08 & \textbf{96$\pm$02} & \underline{88$\pm$04} \\ 
LS & \textbf{89$\pm$17} & 90$\pm$13 & \underline{89$\pm$14} & \textbf{84$\pm$10} & 93$\pm$05 & 88$\pm$06 \\
\rowcolor{gray!10}GB & 52$\pm$17 & \textbf{97$\pm$14} & 67$\pm$15 & 60$\pm$17 & 96$\pm$07 & 72$\pm$10 \\
CV & 65$\pm$26 & 84$\pm$19 & 69$\pm$21 & 47$\pm$19 & 88$\pm$18 & 57$\pm$12 \\
\rowcolor{gray!10}OT & 63$\pm$22 & 72$\pm$25 & 66$\pm$23 & 39$\pm$26 & 42$\pm$31 & 38$\pm$25 \\
VBS & 85$\pm$---  & 86$\pm$---  & -- &  68$\pm$---  & 88$\pm$---  & --\\
 \bottomrule
\end{tabular}
\end{table}

\autoref{tb:AJD2} reports the average Precision (P), Recall (R) and F-Measure (F) achieved by each method in our experiments separated by cases `All', `Normal', `Benign' and `Malignant'.
Note that, for P, R and F: 
\begin{itemize}
    \item BGrowth (BG)  presented the best balance between precision and recall (i.e. F-measure) and the second best results of precision and recall for all cases; 
    \item Growcut (GC)  presented the highest recall and the second best value of F-measure for the `Malignant' case; 
    \item LazySnapping (LS) presented the best results of precision for all cases and the second best values of F-measure for cases `All', `Normal' and `Benign'; 
\end{itemize}

Although GrabCut (GB) presented the best values of recall for most cases, the precision is one of the lowest, which implicates that the method over-segments outside the region of the vertebral body.
LazySnapping (LS) presented the best precision, it means that the method segments a bigger region inside of the vertebral body.
However, for the F-measure LS has not presented the best results.

\begin{figure*}[!bth]
	\centering
		\footnotesize
		\setlength{\tabcolsep}{3pt} 
	\begin{tabular}{cccc}

	\includegraphics[width=0.24\linewidth]{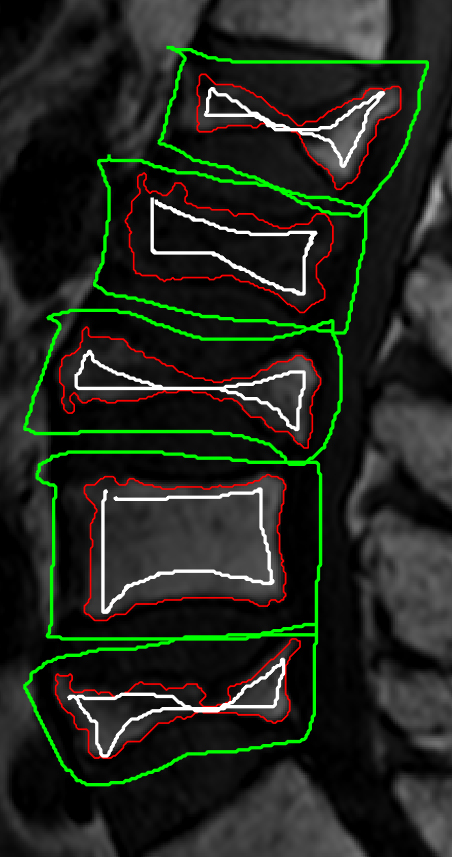} &
	     \includegraphics[width=0.24\linewidth]{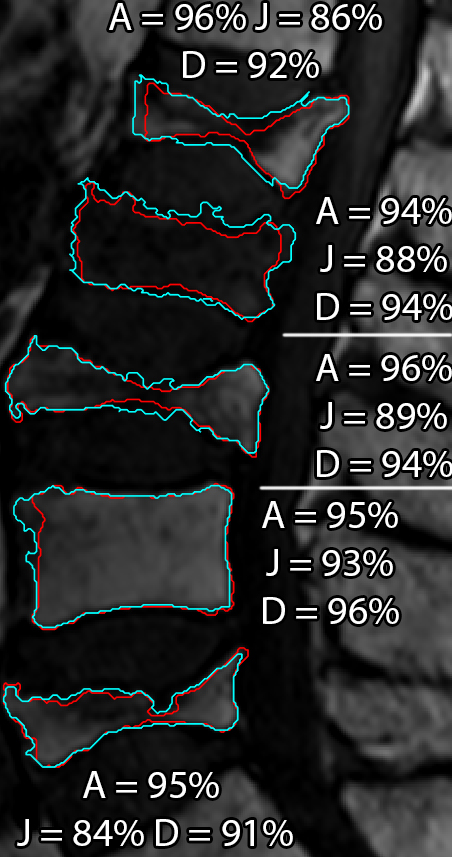} &
	      
	     \includegraphics[width=0.24\linewidth]{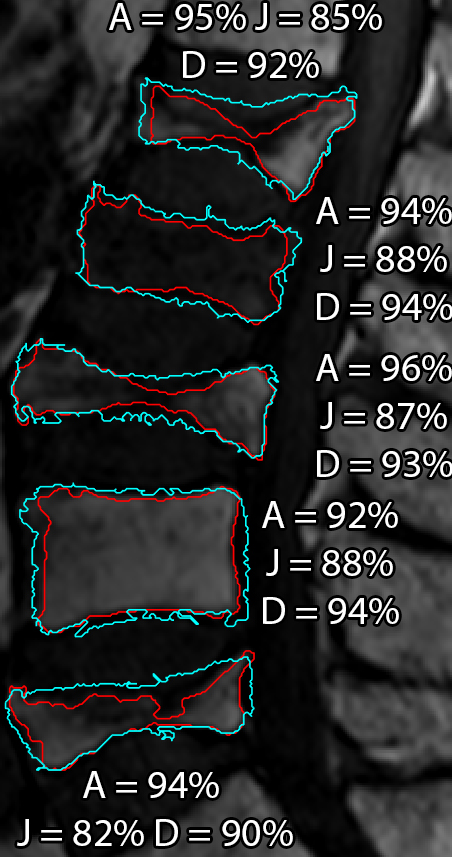} &
	     \includegraphics[width=0.24\linewidth]{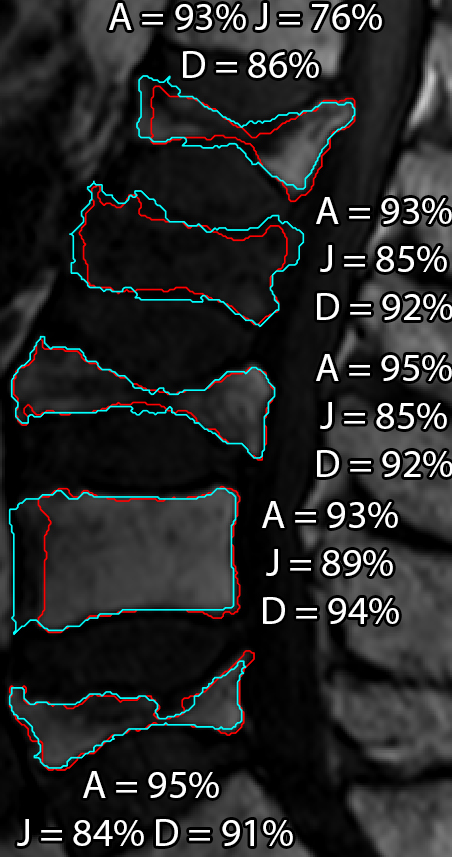} \\ 
	     	(a) GT and Annotations   & (b) BGrowth (BG) & (c) GrowCut (GC) & (d) LazySnapping (LS)\\ 
	\end{tabular}
		\caption{Segmentation results for a case with all lumbar vertebrae with malignant compression fractures. The percentages shown represents the Jaccard Coefficient. Ground-truth in red and semi-automatic segmentation in cyan.}
	\label{fig:segEx}
\end{figure*}

Comparing our proposed BGrowth (BG) to GrowCut (GC), in general, BG keeps similar recall and achieves better precision, consequently, a higher F-measure.
 \autoref{fig:segEx} shows the segmentation results for BGrowth, GrowCut and LazySnapping in a case with three benign VCFs in L1, L3 and L5 (L2 and L4 are normal). 
Note that, the dark and bright regions within the same vertebral bodies difficult the segmentation for all methods, presenting spiked borders due to the neighbours structure intensities, which are very similar (dark).
On these cases, even the manual segmentation is challenging and requires much experience of the human operator.

BG generated smother and more delineated borders, which are closer the ground-truth.
The LS  also produced promising results.
However, sometimes it fails to delimit borders when the interior annotation is closer to the ground-truth and the exterior annotation is not, as shown in results for L4 in \autoref{fig:segEx}.

Analyzing the running time, as \autoref{tb:RunningTime} shows, BGrowth (BG) and GrowCut (GC) presented the same results.
However, BG presented a smaller standard deviation, which indicates BG had less variation in the running time.
Although the Otsu (OT) method presented the fastest running time, in most of the cases, OT presented the lowest values of accuracy (64\% for the malignant case).
Compared to BG and GC, LS is about five times slower, GB is almost four times slower and CV is highly slower (about 30 times).
\begin{table}[!hbt]
\centering
	\footnotesize
		\setlength{\tabcolsep}{3pt} 
\caption{Running time comparison in seconds ($s$).}
\label{tb:RunningTime}
\begin{tabular}{rc}
\toprule
Method & Exam (5 ROIs)/$s$\\ \midrule \midrule
BGrowth (BG) &  \underline{0.154  $\pm$  0.048} \\ 
GrowCut (GC) &   0.154  $\pm$  0.066 \\
LazySnapping (LS) & 0.629  $\pm$  0.301\\
GrabCut (GB)  & 0.492  $\pm$  0.184\\
Chan-Vese (CV) & 3.360  $\pm$  0.360\\
Otsu (OT) & \textbf{0.007 $\pm$   0.011}\\
\bottomrule
\end{tabular}
\end{table}

\subsection{Statistical testing}
As the data presents several identical values, the Kolmogorov-Smirnov~\cite{kolmogorov_1951} test was used for the analysis of the normality of the data.
The null hypothesis that the data follow a normal distribution was rejected for all measures on each segmentation algorithm.
Therefore, the Wilcoxon~\cite{Wilcoxon1970171} test was used to analyze if there are significant differences among the segmentation methods.
In this test, the null hypothesis is that data from two measures, e.g.  the precision from BGrowth (BG) and the precision from GrowCut (GC), are samples from continuous distributions with equal medians, against the alternative that they are not.

\autoref{tb:WilcoxonTest} shows the results for the Wilcoxon test.
Note that, BGrowth (BG) presented significantly better results than GrabCut (GB), Chan-Vese (CV) and Otsu (OT) for each measure for all cases and than LazySnapping (LS) for the recall on every case.
\begin{table}[h]
\centering
\caption{Hypothesis testing results: the result \textcolor{kellygreen}{\checkmark} indicates a rejection of the null hypothesis, i.e. there is significant difference among the same measures for the two methods, and \bm{$\times$} indicates a failure to reject the null hypothesis  (no significant difference) at the 1\% significance level.}
\label{tb:WilcoxonTest}
\footnotesize
\setlength{\tabcolsep}{3pt} 
\begin{tabular}{rcccc}

\toprule
 & \multicolumn{4}{c}{Vertebral bodies considered} \\  \cmidrule{2-5}
 & \multicolumn{1}{c}{All} & \multicolumn{1}{c}{Normal} & \multicolumn{1}{c}{Benign} & \multicolumn{1}{c}{Malignant} \\ 
\multirow{-2}{*}{ \rotatebox[origin=c]{-90}{BG $\times$}} & A J D P R F & A J D P R F & A J D P R F & A J D P R F \\ \midrule \midrule
 GC & \tiny{\textcolor{kellygreen}{\checkmark}} \tiny{\textcolor{kellygreen}{\checkmark}} \tiny{\textcolor{kellygreen}{\checkmark}} \tiny{\textcolor{kellygreen}{\checkmark}} \tiny{\bm{$\times$}} \tiny{\textcolor{kellygreen}{\checkmark}} & \tiny{\textcolor{kellygreen}{\checkmark}} \tiny{\textcolor{kellygreen}{\checkmark}} \tiny{\textcolor{kellygreen}{\checkmark}} \tiny{\textcolor{kellygreen}{\checkmark}} \tiny{\bm{$\times$}} \tiny{\textcolor{kellygreen}{\checkmark}} & \tiny{\textcolor{kellygreen}{\checkmark}} \tiny{\textcolor{kellygreen}{\checkmark}} \tiny{\textcolor{kellygreen}{\checkmark}} \tiny{\textcolor{kellygreen}{\checkmark}} \tiny{\bm{$\times$}} \tiny{\textcolor{kellygreen}{\checkmark}} & \tiny{\bm{$\times$}} \tiny{\bm{$\times$}} \tiny{\bm{$\times$}} \tiny{\bm{$\times$}} \tiny{\bm{$\times$}} \tiny{\bm{$\times$}}\\
 LS & \tiny{\bm{$\times$}} \tiny{\textcolor{kellygreen}{\checkmark}} \tiny{\textcolor{kellygreen}{\checkmark}} \tiny{\textcolor{kellygreen}{\checkmark}} \tiny{\textcolor{kellygreen}{\checkmark}} \tiny{\textcolor{kellygreen}{\checkmark}} & \tiny{\bm{$\times$}} \tiny{\bm{$\times$}} \tiny{\bm{$\times$}} \tiny{\textcolor{kellygreen}{\checkmark}} \tiny{\textcolor{kellygreen}{\checkmark}} \tiny{\bm{$\times$}} & \tiny{\bm{$\times$}} \tiny{\bm{$\times$}} \tiny{\bm{$\times$}} \tiny{\bm{$\times$}} \tiny{\textcolor{kellygreen}{\checkmark}} \tiny{\bm{$\times$}} & \tiny{\bm{$\times$}} \tiny{\bm{$\times$}} \tiny{\bm{$\times$}} \tiny{\bm{$\times$}} \tiny{\textcolor{kellygreen}{\checkmark}} \tiny{\bm{$\times$}}\\ \hline
 GB & \tiny{\textcolor{kellygreen}{\checkmark}} \tiny{\textcolor{kellygreen}{\checkmark}} \tiny{\textcolor{kellygreen}{\checkmark}} \tiny{\textcolor{kellygreen}{\checkmark}} \tiny{\textcolor{kellygreen}{\checkmark}} \tiny{\textcolor{kellygreen}{\checkmark}} & \tiny{\textcolor{kellygreen}{\checkmark}} \tiny{\textcolor{kellygreen}{\checkmark}} \tiny{\textcolor{kellygreen}{\checkmark}} \tiny{\textcolor{kellygreen}{\checkmark}} \tiny{\textcolor{kellygreen}{\checkmark}} \tiny{\textcolor{kellygreen}{\checkmark}} & \tiny{\textcolor{kellygreen}{\checkmark}} \tiny{\textcolor{kellygreen}{\checkmark}} \tiny{\textcolor{kellygreen}{\checkmark}} \tiny{\textcolor{kellygreen}{\checkmark}} \tiny{\textcolor{kellygreen}{\checkmark}} \tiny{\textcolor{kellygreen}{\checkmark}} & \tiny{\textcolor{kellygreen}{\checkmark}} \tiny{\textcolor{kellygreen}{\checkmark}} \tiny{\textcolor{kellygreen}{\checkmark}} \tiny{\textcolor{kellygreen}{\checkmark}} \tiny{\textcolor{kellygreen}{\checkmark}} \tiny{\textcolor{kellygreen}{\checkmark}}\\
 CV & \tiny{\textcolor{kellygreen}{\checkmark}} \tiny{\textcolor{kellygreen}{\checkmark}} \tiny{\textcolor{kellygreen}{\checkmark}} \tiny{\textcolor{kellygreen}{\checkmark}} \tiny{\textcolor{kellygreen}{\checkmark}} \tiny{\textcolor{kellygreen}{\checkmark}} & \tiny{\textcolor{kellygreen}{\checkmark}} \tiny{\textcolor{kellygreen}{\checkmark}} \tiny{\textcolor{kellygreen}{\checkmark}} \tiny{\textcolor{kellygreen}{\checkmark}} \tiny{\textcolor{kellygreen}{\checkmark}} \tiny{\textcolor{kellygreen}{\checkmark}} & \tiny{\textcolor{kellygreen}{\checkmark}} \tiny{\textcolor{kellygreen}{\checkmark}} \tiny{\textcolor{kellygreen}{\checkmark}} \tiny{\textcolor{kellygreen}{\checkmark}} \tiny{\textcolor{kellygreen}{\checkmark}} \tiny{\textcolor{kellygreen}{\checkmark}} & \tiny{\textcolor{kellygreen}{\checkmark}} \tiny{\textcolor{kellygreen}{\checkmark}} \tiny{\textcolor{kellygreen}{\checkmark}} \tiny{\textcolor{kellygreen}{\checkmark}} \tiny{\textcolor{kellygreen}{\checkmark}} \tiny{\textcolor{kellygreen}{\checkmark}}\\
 OT & \tiny{\textcolor{kellygreen}{\checkmark}} \tiny{\textcolor{kellygreen}{\checkmark}} \tiny{\textcolor{kellygreen}{\checkmark}} \tiny{\textcolor{kellygreen}{\checkmark}} \tiny{\textcolor{kellygreen}{\checkmark}} \tiny{\textcolor{kellygreen}{\checkmark}} & \tiny{\textcolor{kellygreen}{\checkmark}} \tiny{\textcolor{kellygreen}{\checkmark}} \tiny{\textcolor{kellygreen}{\checkmark}} \tiny{\textcolor{kellygreen}{\checkmark}} \tiny{\textcolor{kellygreen}{\checkmark}} \tiny{\textcolor{kellygreen}{\checkmark}} & \tiny{\textcolor{kellygreen}{\checkmark}} \tiny{\textcolor{kellygreen}{\checkmark}} \tiny{\textcolor{kellygreen}{\checkmark}} \tiny{\textcolor{kellygreen}{\checkmark}} \tiny{\textcolor{kellygreen}{\checkmark}} \tiny{\textcolor{kellygreen}{\checkmark}} & \tiny{\textcolor{kellygreen}{\checkmark}} \tiny{\textcolor{kellygreen}{\checkmark}} \tiny{\textcolor{kellygreen}{\checkmark}} \tiny{\textcolor{kellygreen}{\checkmark}} \tiny{\textcolor{kellygreen}{\checkmark}} \tiny{\textcolor{kellygreen}{\checkmark}}\\
\bottomrule
\end{tabular}
\end{table}
LS presented significantly better precision than BGrowth for cases `All' and `Normal'.
On the other hand, on case `All', BGrowth presented significantly better results than LazySnapping for Jaccard, Dice and F-measure.
BGrowth improved GrowCut's performance in the three first cases (`All', `Normal' and `Benign'), for all measures (except the recall, in which BG kept performance similar to GC).
On the case `Malignant', BGrowth kept  similar results to GrowCut for every measure.

\subsection{Variation on the annotations}

To assess the effect of the details employed on the interior and exterior manual annotations, we conducted two different analysis, which are detailed as follows: 
\begin{enumerate}
    \item We vary the percentage of interior annotation on each the vertebral body, and keep a sloppy exterior annotation done (as previous shown in \autoref{fig:an-examples}).
    For each vertebral body, a percentage of the ground-truth is extracted, always keeping at least a line that goes trough the dark and bright regions. 
    The percentage starts at 10\%, increasing by 10\%, up to 100\% (the ground-truth itself), summing up to 10 variations.
    
    \item We vary the distance in pixels of the exterior annotation in relation to the ground-truth and keep a sloppy interior annotation.
    Starting at a distance of 3 pixels from the ground-truth, increasing by 3 up to 30 pixels, which sums up to 10 variations.
    If the distance surpass the image grid, the last pixel is assigned as boundary.
\end{enumerate}

\begin{figure}[!bth]
	\centering
	\centering	
\footnotesize
\setlength{\tabcolsep}{3pt} 
	\begin{tabular}{ccc}
	\multicolumn{3}{c}{\textbf{Interior Annotation}} \\ 
	     10\% & 50\% & 80\% \\
	     \includegraphics[width=0.3\linewidth]{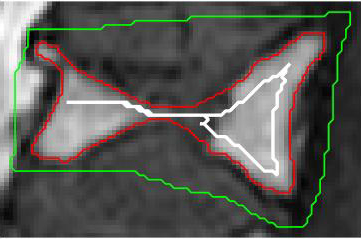} &
	     \includegraphics[width=0.3\linewidth]{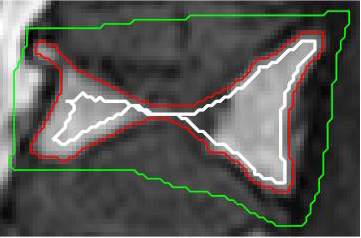} &
	     \includegraphics[width=0.3\linewidth]{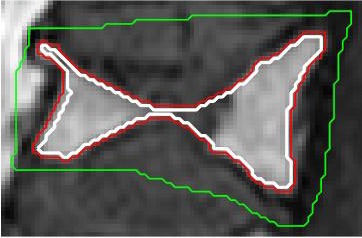} \\
	      \multicolumn{3}{c}{\textbf{Exterior Annotation}} \\ 
	     3 pixels & 15 pixels & 30 pixels \\
	     \includegraphics[width=0.3\linewidth]{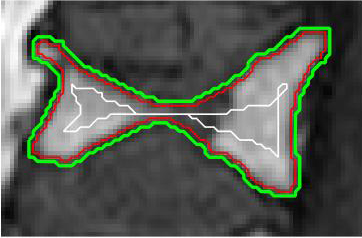} & 
	     \includegraphics[width=0.3\linewidth]{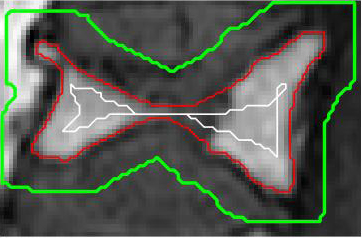} &
	     \includegraphics[width=0.3\linewidth]{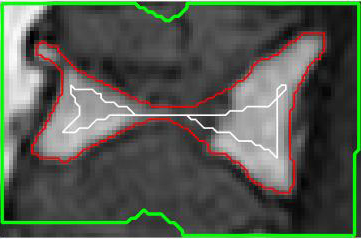}\\ 
	     
	\end{tabular}
		\caption{Examples of interior and exterior regions used.}
	\label{fig:InOutAnnotation}
\end{figure}

\begin{figure}[!bth]
	\centering
	\centering	
\footnotesize
\setlength{\tabcolsep}{3pt} 
	\includegraphics[width=0.8\linewidth]{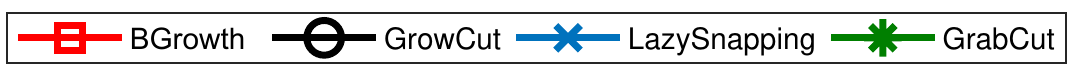}
	\begin{tabular}{c}
	\includegraphics[width=0.45\linewidth]{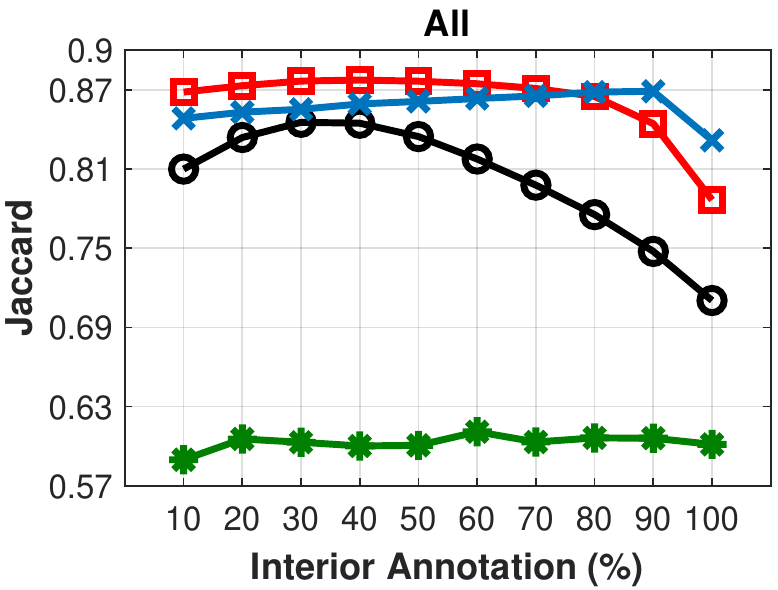} 
	\includegraphics[width=0.45\linewidth]{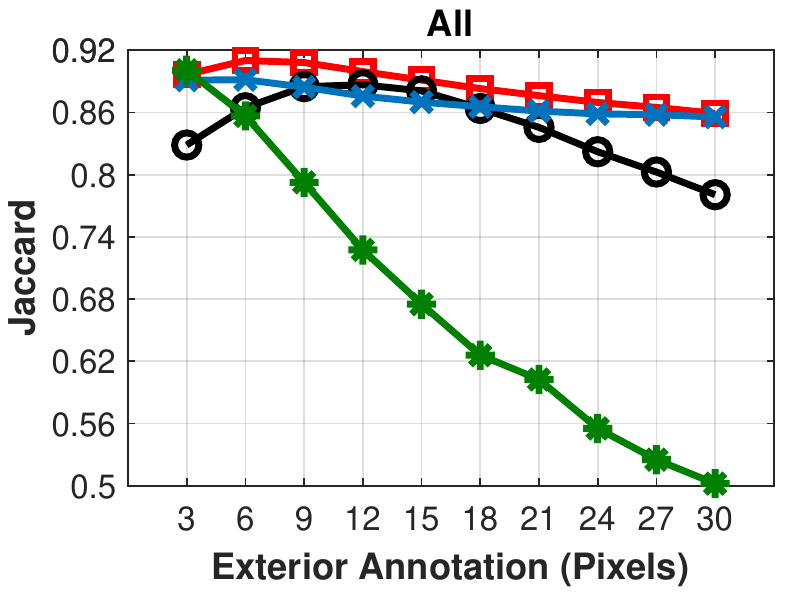} \\
	     \includegraphics[width=0.45\linewidth]{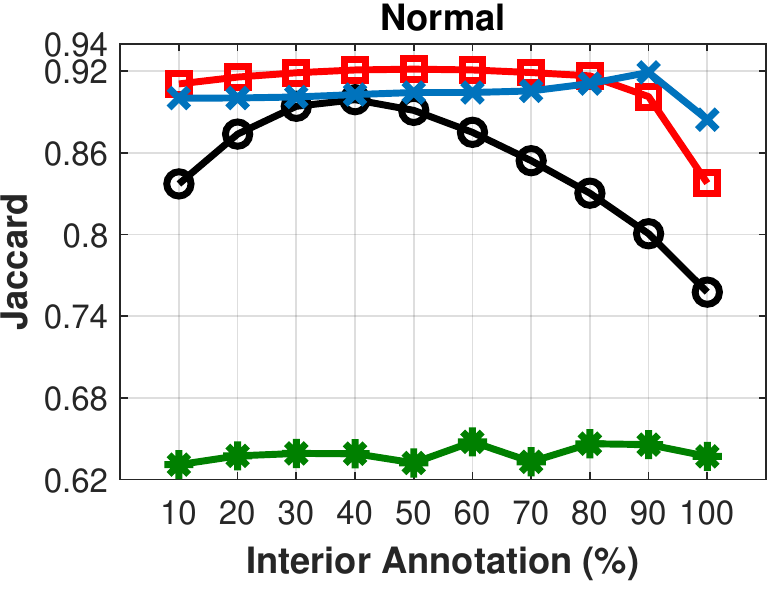} 
	     \includegraphics[width=0.45\linewidth]{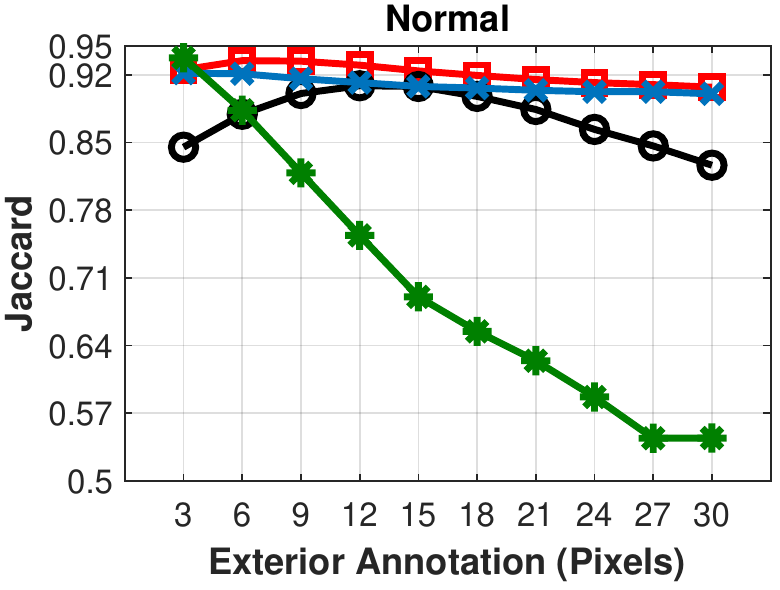}  \\
	     \includegraphics[width=0.45\linewidth]{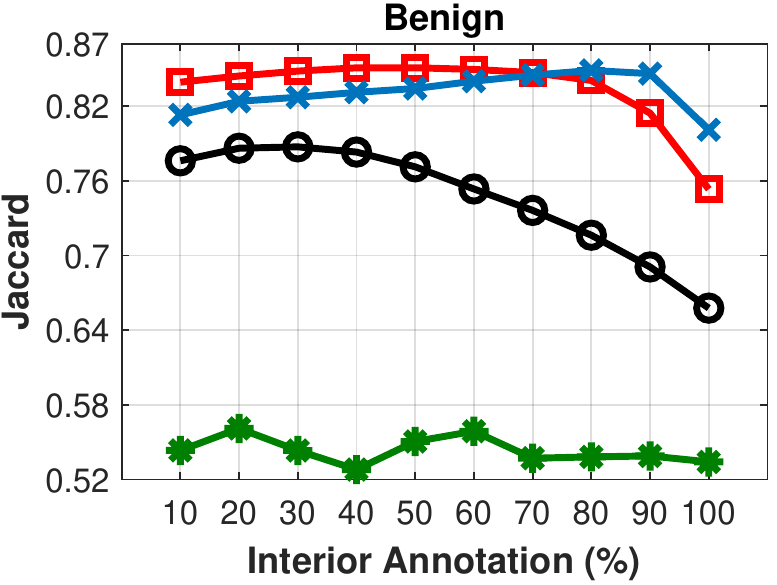} 
	     \includegraphics[width=0.45\linewidth]{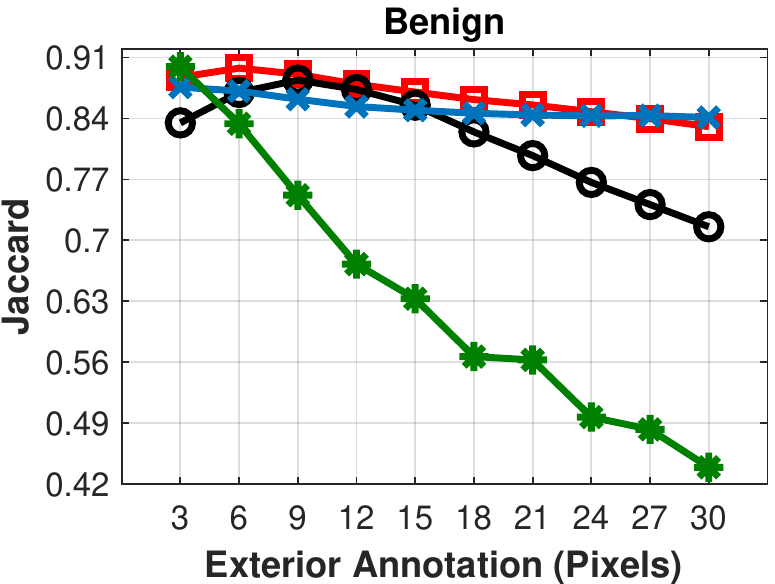} \\
	     \includegraphics[width=0.45\linewidth]{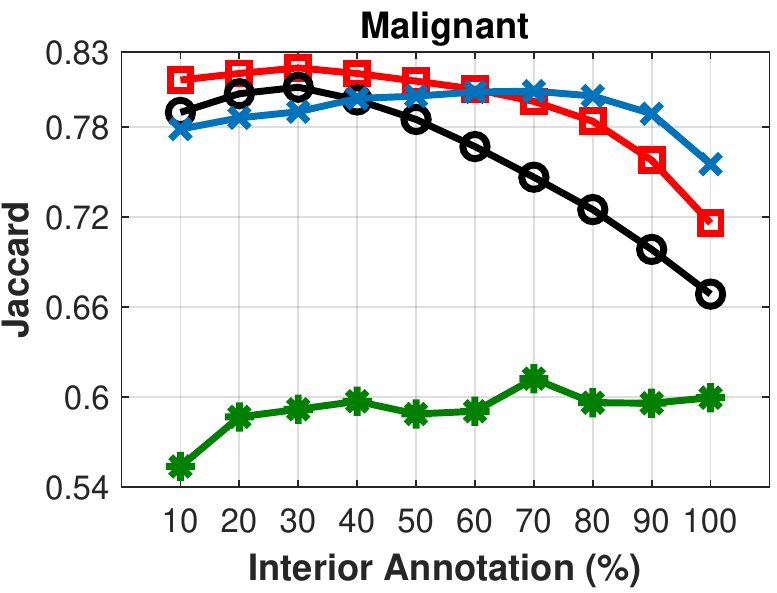}  
	     \includegraphics[width=0.45\linewidth]{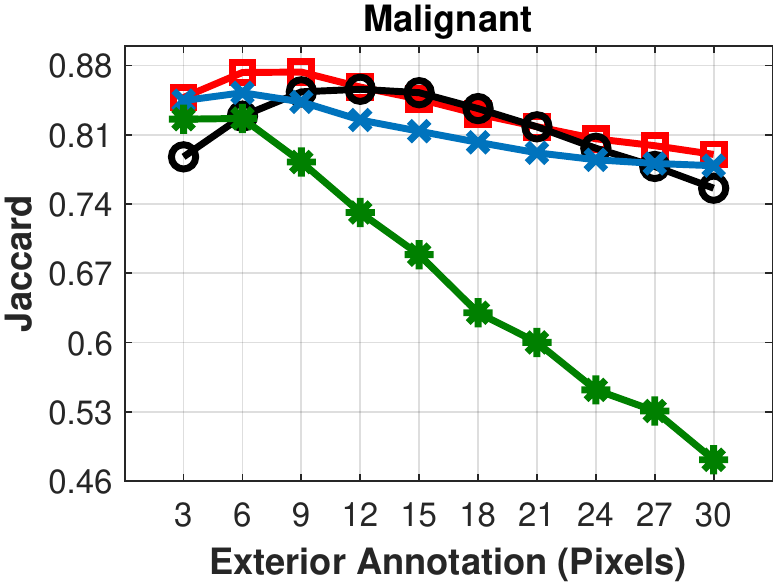} \\ 
	     
	\end{tabular}
		\caption{Comparison of Jaccard coefficient over annotation variations at the interior and exterior of each vertebral body.}
	\label{fig:inOutCasesVariation}
\end{figure}

\autoref{fig:InOutAnnotation} shows a few examples of variations of the interior percentages and the distances from the ground-truth used.
\autoref{fig:inOutCasesVariation} shows the results for the interior and exterior variations for each one of the four cases using the Jaccard measure.
Note that, for the interior annotation variation, BGrowth (BG) leads with the best results for all cases up to 60\%.
Besides, for all percentages and all cases, BG presented better results than GrowCut (GC).
GrabCut (GB) presented the lowest results while LazySnapping (LS) presented the best results from 90\% to 100\%.
However, in a real case scenario, 90\% or  100\% interior annotation rarely happens\footnote{It is practically the ground-truth itself, which takes too much time to annotate.}.

For the exterior annotation variation, GrabCut (GB) drops the performance really fast for all cases.
GrowCut (GC) achieves its peak at 12 pixels for all cases, and then dropped the performance slowly.
BGrowth (BG) presented better or similar performance than GC for all measures in all cases.
LazySnapping (LS) presented results closer to BG with a higher difference from 6 to 18 pixels pixels for all cases.
In general BG presented the best Jaccard values with ``sloppy'' annotation inside the vertebral body, which is very helpful, as the specialist does not need to spend much time drawing detailed curves at the interior region.
The same holds for the exterior region annotation.

\begin{figure}[!bth]
	\centering
	\centering	
\footnotesize
\setlength{\tabcolsep}{3pt} 
	\includegraphics[width=0.8\linewidth]{images/myLegend.pdf}
	
	\begin{tabular}{cc}
	\includegraphics[width=0.45\linewidth]{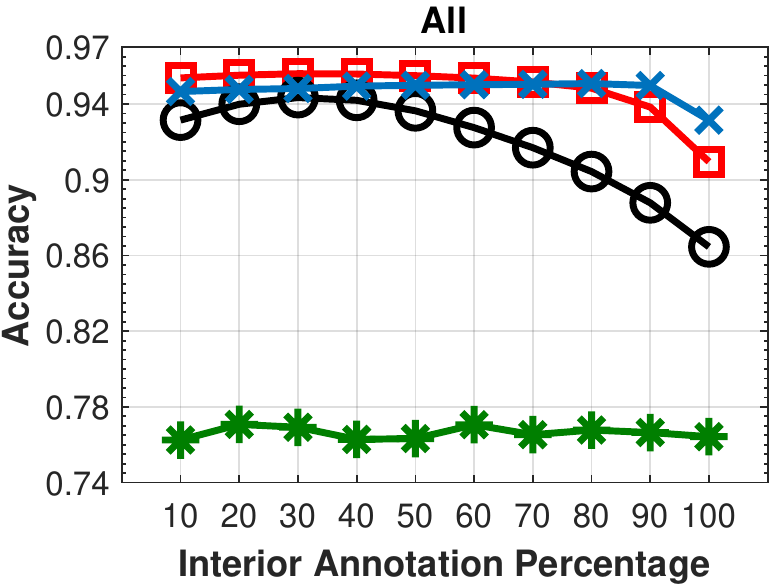}  &
	     \includegraphics[width=0.45\linewidth]{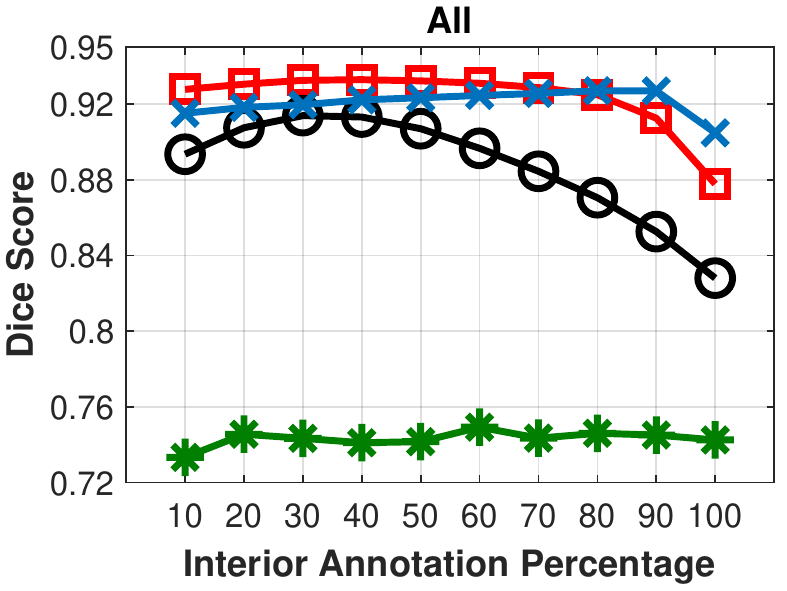} \\  
	     \includegraphics[width=0.45\linewidth]{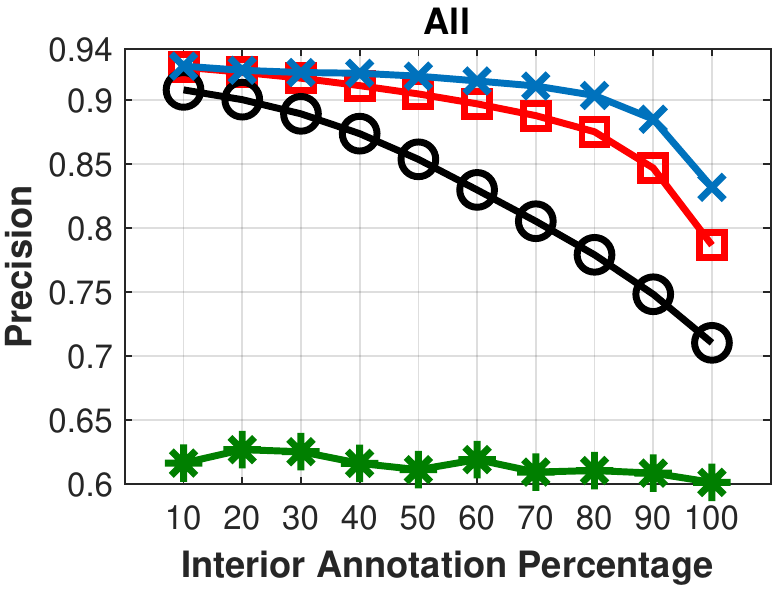}  &
	     \includegraphics[width=0.45\linewidth]{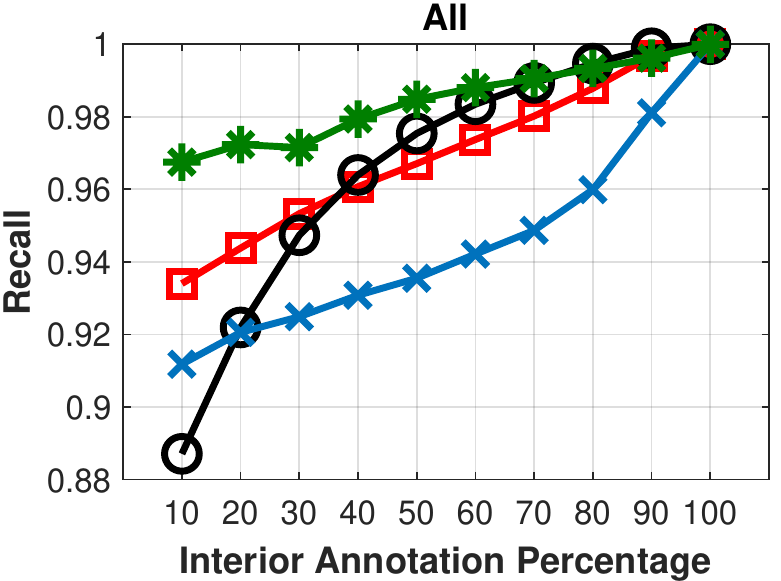} \\
        \includegraphics[width=0.45\linewidth]{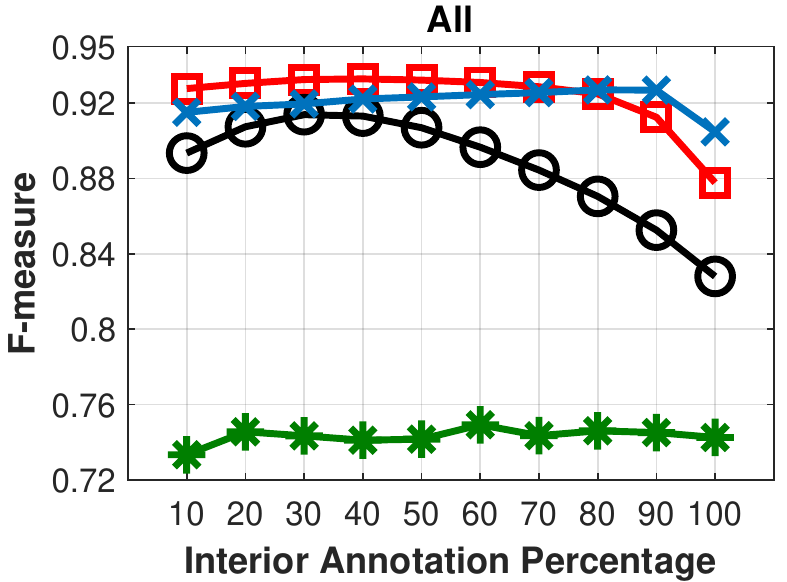}  &
\includegraphics[width=0.45\linewidth]{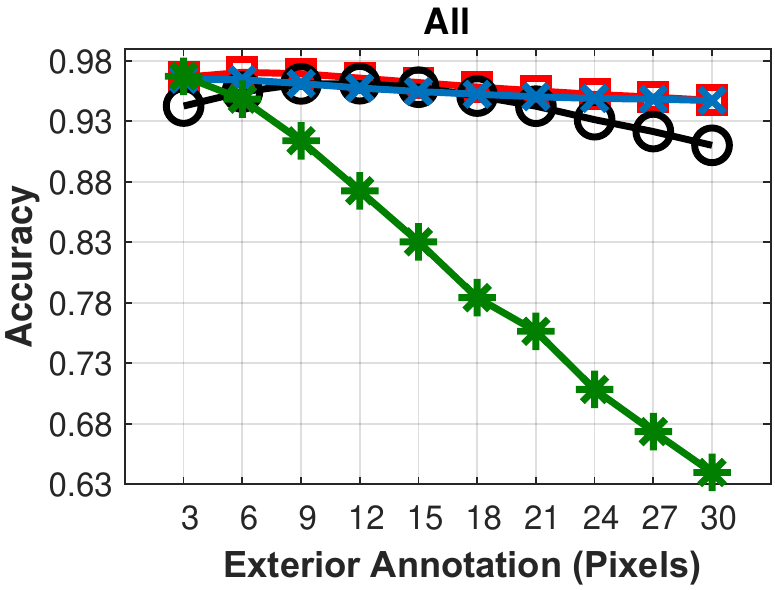} \\
 \includegraphics[width=0.45\linewidth]{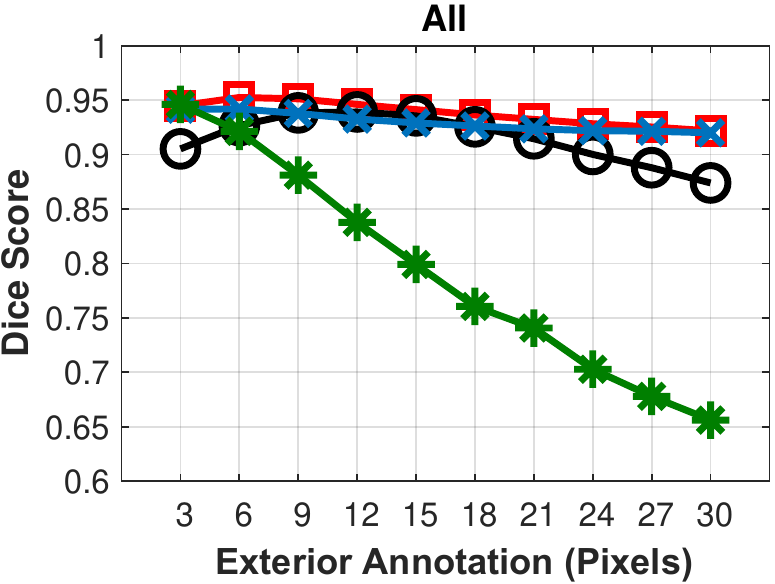} &
 \includegraphics[width=0.45\linewidth]{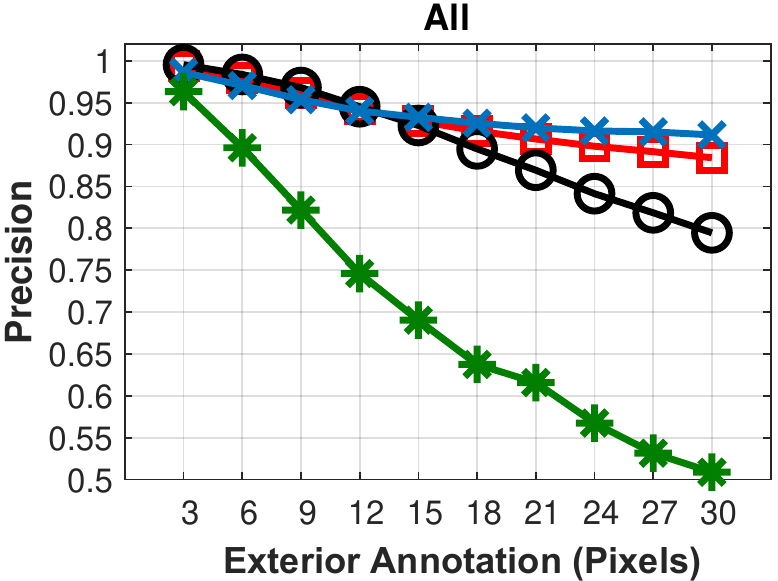}  \\
  \includegraphics[width=0.45\linewidth]{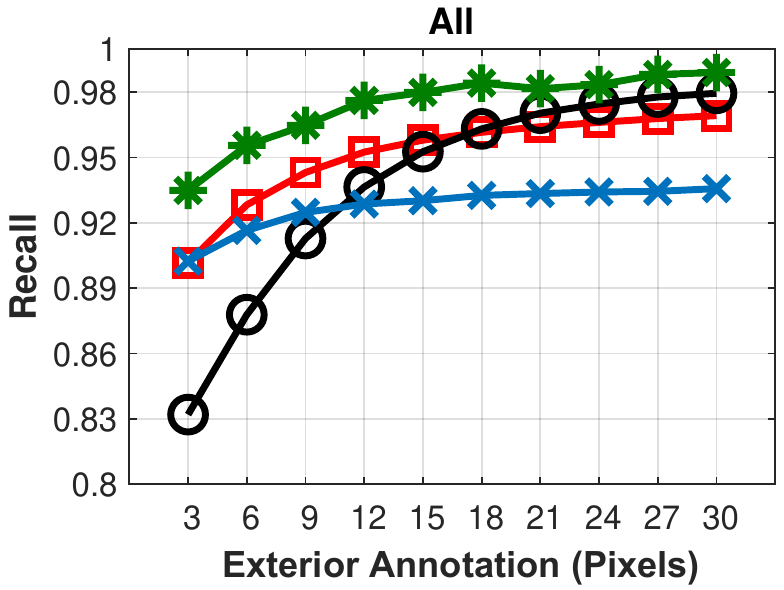} &
  \includegraphics[width=0.45\linewidth]{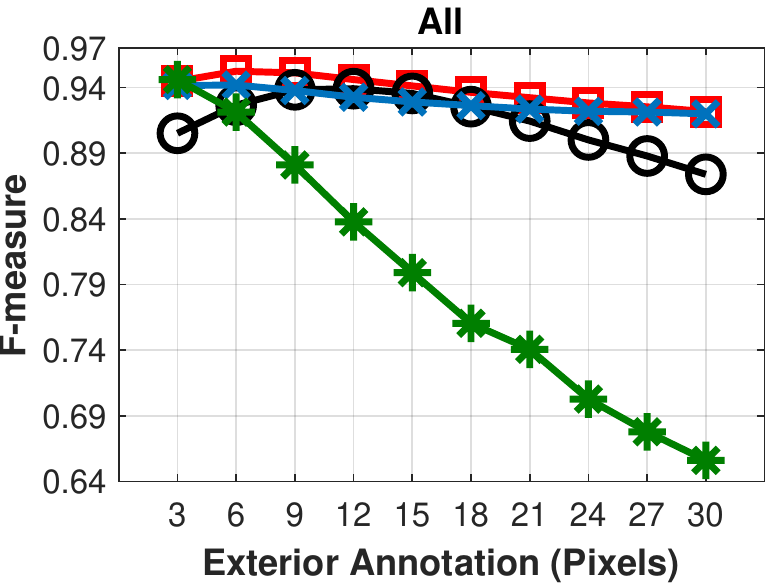} 
	     
	\end{tabular}
		\caption{Comparison of several measures over annotation variations at the interior and exterior of each vertebral body.}
	\label{fig:outsideVariation}
\end{figure}

To further investigate, \autoref{fig:outsideVariation} shows the results for the other five measures for the general case (`All').
Note that, for the interior annotation variation, BGrowth (BG) only presented lower results than GrowCut (GC) for the recall from 40\% to 80\% of interior annotation.  
However, the F-measure is higher for BG, which implicates that, in general, BG presented a better balance between precision and recall than GC.
Although LazySnapping (LS) presented higher precision most of the time, its recall is one of the lowest.
On the other hand, GrabCut (GB) presented one of the highest recall and lower values for the other measures.

Considering the interior annotation variation, for the accuracy, Dice, precision and F-measure, GB drops fast as the distance from the ground-truth boundary increases.
GC, BG and LS presented quite similar behavior from 3 to 12 pixels.
Over 12 pixels, there are small differences, in which GC presented the lowest results and BG and LS presented almost the same results.

In general, in a real case scenario, BGrowth (BG) is faster (as shown in~\autoref{tb:RunningTime}) and produces better or similar results than LS even with sloppy interior and exterior annotations. Note that LS demands a more precise annotation in order to achieve better results, demanding more effort from the specialist.

\section{Conclusions and future works}
\label{sec:conclusion}

The semi-automatic segmentation of Vertebral Compression Fractures (VCFs) is a challenging task: in most cases, they present several regions with non-homogeneous intensities within the same vertebral body.
We have investigated this challenge and we proposed an efficient and accurate method called BGrowth (BG), which balances the weights of the regions in expansion.
The segmentation performance obtained by BGrowth significantly outperforms other well-known methods from the literature and keeps an equivalent running time regarding the fastest competitors.
Balanced Growth presents the best results with sloppy annotations, which demands less effort of the specialist on marking seeds points.

BGrowth presented an accuracy of 94\% while GrowCut and LazySnapping presented 93\% and 94\%, respectively.
Although LazySnapping presented results closer to BGrowth, its running time is almost five times slower.

As future work, we intend to improve BGrowth segmentation results by dynamically adapting the weights during the region expansion process and make it even easier for the specialist to annotate the seed points.
Moreover, the segmentation results achieved can be used for feature extraction and classification.

\section*{Acknowledgments}
This work is supported by the S\~ao Paulo Research Foundation (FAPESP, grants No. 2017/23780-2, 2016/17078-0) and by the Coordination for the Improvement of Higher Level -or Education- Personnel (CAPES, grant No.: 0487/17083480), and National Council for Scientific and Technological Development (CNPq).

\bibliographystyle{abntex2-num}
\small{\bibliography{referencia}}

\end{document}